\documentclass[letterpaper, 10 pt, conference]{ieeeconf}  

\IEEEoverridecommandlockouts                              

\usepackage[dvipsnames]{xcolor}
\usepackage{graphicx}
\usepackage{paralist}
\usepackage{subcaption}
\usepackage{booktabs}

\usepackage{tikz}
\usepackage{comment}
\usepackage{amsmath,amssymb} 
\usepackage{color}
\usepackage{xcolor}
\usepackage{colortbl}
\usepackage{multirow}
\usepackage{pifont}
\newcommand{\cmark}{\ding{51}}%
\newcommand{\xmark}{\ding{55}}%
\definecolor{RowColorCode}{rgb}{0.61,0.57,0.89}

\title{\LARGE \bf
Differentiable Frequency-based Disentanglement for \\ Aerial Video Action Recognition
}

\author{Divya Kothandaraman, Ming Lin and Dinesh Manocha \\ University of Maryland College Park \\ https://gamma.umd.edu/diffar
}

\begin{document}

\maketitle
\thispagestyle{empty}
\pagestyle{empty}


\begin{abstract}

We present a learning algorithm for human activity recognition in videos. Our approach is designed for UAV videos, which are mainly acquired from obliquely placed dynamic cameras that contain a human actor along with background motion. Typically, the human actors occupy less than one-tenth of the spatial resolution. Our approach simultaneously harnesses the benefits of frequency domain representations, a classical analysis tool in signal processing, and data driven neural networks. We build a differentiable static-dynamic frequency mask prior to model the salient static and dynamic pixels in the video, crucial for the underlying task of action recognition. We use this differentiable mask prior to enable the neural network to intrinsically learn disentangled feature representations via an identity loss function. Our formulation empowers the network to inherently compute disentangled salient features within its layers. Further, we propose a cost-function encapsulating temporal relevance and spatial content to sample the most important frame within uniformly spaced video segments. We conduct extensive experiments on the UAV Human dataset and the NEC Drone dataset and demonstrate relative improvements of $5.72\% - 13.00\%$ over the state-of-the-art and $14.28\% - 38.05\%$ over the corresponding baseline model.

\end{abstract}

\section{Introduction}

UAVs are being increasingly used for surveillance, security, inspection, search and rescue, agriculture and videography. Action recognition \cite{rodriguez2019video} for videos captured by these UAVs is a crucial video analysis task. While there has been immense work on videos obtained using front-view cameras, research on UAV videos is limited. Deep neural networks for human activity recognition~\cite{feichtenhofer2019slowfast, feichtenhofer2020x3d} have been widely used for ground-camera scenes~\cite{carreira2017quo,monfort2019moments}, where the human actors take a high fraction of the pixels in the video scenes. 
While these methods work well on ground-camera videos, their accuracy on UAV videos can be low. This is mainly because UAV videos consist of human actors that occupy less than $10\%$ of the spatial resolution of the video. Moreover, the action captured by these cameras may be low-resolution, and there is background and camera movement. 

\begin{figure}[h]
    \centering
    \includegraphics[scale=0.32]{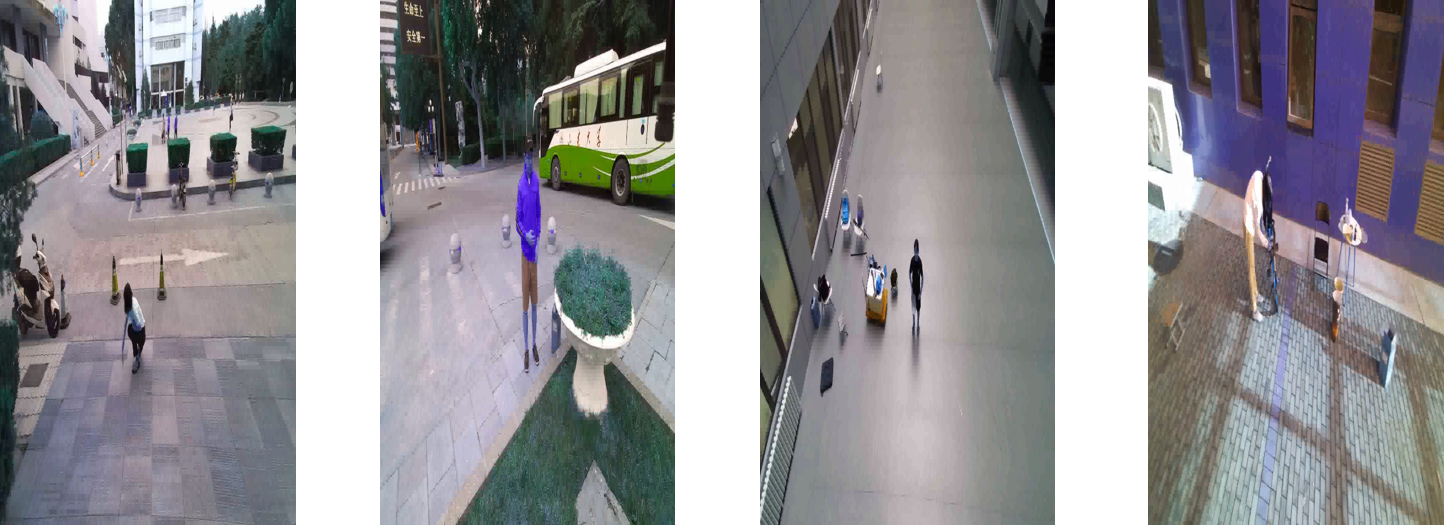}
    \caption{Human activity recognition in videos recorded using dynamic cameras at oblique and aerial angles typically consist of human actors that occupy less than $10\%$ of the spatial resolution of the video. Background motion, noise, and low-resolution activities add to the complexities faced by video recognition systems. Our method achieves upto $41.86\%$ and $80.53\%$ and top-1 accuracies on UAV Human and NEC Drone, respectively.}
    \label{fig:coverPic}
\end{figure}

Moving objects in videos can be detected using object detection and tracking~\cite{ren2015faster}, optical flow~\cite{dosovitskiy2015flownet}, motion segmentation~\cite{zappella2008motion}, etc. However, many issues arise in terms of using them on UAV videos: unavailability of ground-truth object detection labels, domain gap between scenes with access to object localization and scenes on which action recognition needs to be performed~\cite{wang2018deep,chen2018domain,benjdira2019unsupervised}, the need for a separate learning technique for modeling motion~\cite{gammulle2017two}, etc. Thus, the network needs to automatically disentangle regions of the feature map, which are vital for video recognition, from entangled representations encompassing all aspects of the video. 
 

The design of a differentiable method to compute disentangled representations enables the neural network to intrinsically learn disentangled regions, from motion cues and learned saliency information. A differentiable formulation between the neural network and motion representations establishes a bridge between classical and learning-based techniques, harnessing the power of both. Differentiable programming has been used in diverse tasks such as simulation~\cite{qiao2021differentiable} and control~\cite{takahashi2021differentiable}. However, it has not been explored in action recognition literature.  

\subsection{Main Contributions}

We present a novel method for activity recognition in UAV videos with dynamic backgrounds and moving cameras. Our approach is designed to handle {\em oblique} and {\em aerial} camera views with {\em low actor-background pixel ratios}, which are common characteristics in videos captured using aerial cameras. We take advantage of frequency domain representations \cite{kothandaraman2022fourier} of the feature maps that provide an alternate complementary formulation to holistically model the time-evolution of pixels corresponding to various entities in the scene. We propose a differentiable static-dynamic frequency mask that (i) empowers the network to learn disentangled feature maps via a differentiable loss function and (ii) samples crucial frames from the video at test time. The novel components of our work include:

\begin{enumerate}
    \item {\bf Differentiable static-dynamic mask:} ``Dynamic'' saliency mask denotes important regions of the scene corresponding to moving pixels. Similarly, ``static'' saliency mask represents important regions of the scene corresponding to stationary pixels. We propose a {\em differentiable} frequency-based ``static'' saliency mask that complements the frequency-based ``dynamic'' saliency mask. While the ``dynamic'' mask highlights dynamic salient regions of the scene and suppresses static non-salient regions, our ``static'' mask amplifies static salient regions and suppresses dynamic non-salient regions. Our approach uses a combination of {\em both masks to handle dynamic and static salient regions} of the scene (including {\em moving human actors}) and suppresses dynamic backgrounds, noise, and static non-salient regions. 
    \item {\bf Differentiable disentanglement:} We bestow the network with the ability to inherently ``learn'' disentangled feature representations that are crucial for the underlying task of action recognition, via an identity loss function using the {\em ``differentiable'' static-dynamic frequency mask}. This empowers layers of the network to serve as the disentangler and directly compute salient features in its forward pass.
    \item {\bf Mask sampling:} We propose a frame sampling method using our {\em static-dynamic frequency mask}, to select the most informative frame within uniformly spaced non-overlapping video segments. These frames serve as a complementary descriptor to frames chosen using uniform sampling.  
\end{enumerate}

Our method can be embedded within any 3D action recognition backbone. It is parameter-free i.e. it does not add any learnable layers or parameters to the backbone neural network, which simplifies initialization and training. We conduct extensive experiments on two UAV datasets: UAV Human~\cite{li2021uav} and NEC Drone~\cite{choi2020unsupervised}. These datasets are captured at varying altitudes, angles and lighting conditions. We demonstrate a considerable improvement of {\bf 2.84}$\%$ -- {\bf 13.05}$\%$ in top-1 accuracy over the corresponding baselines. We demonstrate state-of-the-art performance on UAV Human and NEC Drone by at least {\bf 3.26\%} and {\bf 7.74\%}, respectively, over the state-of-the-art methods. Moreover, our method is on par with the state-of-the-art transformer architectures while using far fewer computations (up to {\bf 61.8}$\times$) and network parameters (up to {\bf 31.94}$\times$). 

\section{Related Work}

\paragraph{Aerial Video Recognition.} The availability of UAV datasets~\cite{li2021uav,zhu2018visdrone,bozcan2020air} has fostered aerial video research~\cite{nguyen2022state} pertaining to person reidentification~\cite{fysh2018person}, human detection~\cite{portmann2014people}, tracking~\cite{he2020towards, mueller2016persistent}, pose estimation~\cite{benini2016real}, few-shot learning~\cite{sultani2021human}, drone detection~\cite{ashraf2021dogfight} and path planning~\cite{hayat2017multi}. Many architectures have been proposed to specifically tackle aerial video action recognition~\cite{kotecha2021background,peng2020fully,ding2020lightweight,ulhaq2016action}, in addition to generic action recognition ~\cite{feichtenhofer2019slowfast,feichtenhofer2020x3d}. Recently, FAR~\cite{kothandaraman2022fourier} proposed a frequency-based method to disentangle moving objects by modulating feature maps. However, FAR does not place any explicit emphasis on static regions of the scene, a complementary descriptor. Moreover, FAR applies the mask as a transformation over feature maps, denying the network the opportunity to directly `learn' disentangled feature maps in a differentiable manner.

\paragraph{Differentiability and Deep Learning.} Differentiability has emerged as a powerful tool towards making neural networks learn tailored feature representations for diverse tasks such as simulation~\cite{qiao2021differentiable}, control~\cite{takahashi2021differentiable}, video structure from motion~\cite{teed2018deepv2d}, and physics~\cite{degrave2019differentiable}. In this paper, we present a differentiable frequency-domain method for aerial video action recognition. 

\paragraph{Motion Representation Methods.} A popular motion representation method is optical flow~\cite{beauchemin1995computation}, which is very expensive. Deep learning based optical flow computation~\cite{ren2017unsupervised,ilg2017flownet,dosovitskiy2015flownet} has a lower computational complexity, but requires a two-stream NN~\cite{simonyan2014two} for action recognition that increases the network complexity. Object tracking methods~\cite{zhang2021recent} require ground-truth bounding boxes that are difficult to obtain. Other alternatives such as motion feature networks~\cite{lee2018motion} and ActionFlowNet~\cite{ng2018actionflownet} are less accurate optical flow. Techniques such as background subtraction~\cite{piccardi2004background} and motion segmentation~\cite{zappella2008motion} are not ideal for action recognition methods that rely heavily on object localization~\cite{sengupta2020background,ellenfeld2021deep}. In contrast, we present a computationally efficient frequency-based disentanglement technique to automatically highlight regions of the scene salient for action recognition.

\paragraph{Frame sampling.} Frame sampling~\cite{wu2019adaframe} has mainly been explored in the context of untrimmed videos~\cite{wu2019multi} using Reinforcement learning~\cite{gowda2021smart}, saliency information~\cite{korbar2019scsampler}, audio information~\cite{gao2020listen}, etc. Most of these methods require the neural network to be retrained using the selected frames which is expensive. Moreover, CNN methods such as MG Sampler~\cite{zhi2021mgsampler} use only prior frames, while it is beneficial to use future frames as well in a non-online setting.

\section{Method}

\begin{figure*}
    \centering
    \includegraphics[scale=0.32]{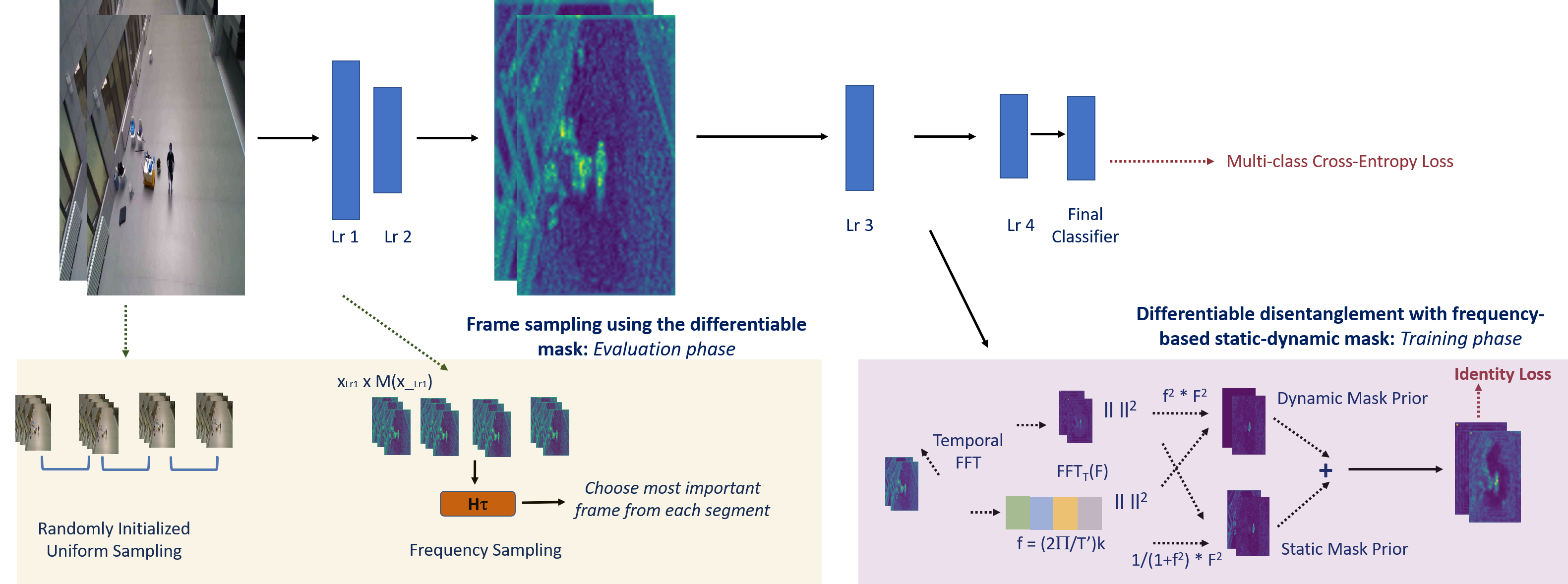}
    \caption{We present a differentiable frequency-based method for aerial video recognition. Our differentiable static-dynamic frequency mask provides a prior for disentangled regions relevant to action recognition. This mask is used to guide the learning of disentangled features within the layers of the neural network using an identity function. Further, we propose a frame sampling strategy that chooses the best frame within each uniform video segment, at test time, using the static-dynamic frequency mask and temporal difference.}
    \label{fig:overview}
\end{figure*}
We present a novel method for video recognition in UAV videos. These videos, recorded using dynamic cameras placed at oblique angles obstructing the field of view of the action being performed, contain human actors occupying fewer than $10\%$ of the pixels, along  with background motion and light noise. Our method has multiple benefits: (i) It can be embedded within any 3D action recognition backbone without modifying the structure of the underlying neural network. (ii) It is parameter-free, i.e., it does not add any learnable layers or parameters to the backbone neural network. These properties allow us to initialize the neural network with weights pre-trained on large-scale ground-camera datasets such as Kinetics~\cite{carreira2017quo} and simplifies training.

\begin{table}
\centering 
\resizebox{0.6\columnwidth}{!}{
\begin{tabular}{cc}
\toprule
Notation & Term   \\
\midrule 
$M_{dynamic}$ & Dynamic mask \\
$M_{static}$ & Static mask \\
$M$ & $M_{dynamic}$ + $M_{static}$ \\
$f$ & Frequencies \\
$F$ & Fourier transform \\
$x$ & Feature representations \\
$C_{p}$ & Prior frames cost \\
$C_{f}$ & Future frames cost \\
$T$ & Frame sampling rate \\
\bottomrule
\end{tabular}
}
\caption{Symbols used in the paper}
\label{tab:notations}
\end{table}

We present our method in Figure \ref{fig:overview}. We sample $T$ frames from the video using uniform sampling~\cite{kothandaraman2022fourier}, where $T$ is the frame rate. The architecture involves a 3D CNN backbone such as I3D~\cite{carreira2017quo} or X3D~\cite{feichtenhofer2020x3d}. 3D CNNs typically consist of an initial set of convolution layers, followed by $4$ blocks of ResNet-type convolution layers. This is followed by a final prediction layer. Our architecture retains the structure of this backbone. We propose a differentiable static-dynamic frequency mask to highlight regions of the scene relevant for action prediction. This is computed using the outputs at the second block of the neural network. The neural network is trained using traditional multi-class cross entropy loss for action recognition and an identity loss. The identity loss function is applied at the outputs of the third block of the neural network, and bestows the network with the ability to learn disentangled feature maps in a differentiable manner via an identity loss function. We also propose a method for frame sampling using the differentiable static-dynamic frequency mask, which samples frames from the video at the evaluation phase. We now describe our method in detail.

\subsection{Background: Differentiability} 

Differentiable programming allows for computing derivatives of functions or gradient operators to learn representations for the underlying task. For feature disentanglement, differentiable programming can be a more powerful tool than applying transformations over features~\cite{kothandaraman2022fourier}. This is because blocks of the network are optimized to directly compute disentangled features, where the entire set of layers serves as the disentanglement module.

Frequency domain representations contain an abundance of information for disentanglement that can improve the performance of action recognition neural networks. One way to do so is to apply the corresponding frequency-domain masks as a transformation over neural network features. However, enabling the neural network to learn disentangled representations within its layers, using frequency-domain information as well as training data, creates an amalgamation of techniques from traditional signal processing and deep learning. This has the potential to empower the neural network with the benefits of both worlds.

However, the training of neural networks requires loss functions to be differentiable in order to perform backpropagation. This raises questions pertaining to frequency domain representations that can be used for training, appropriate loss function and the layers of the neural network that should be used for learning robust disentangled features.

\subsection{Differentiable Static-Dynamic Frequency Mask}

Consider a video depicting the action of human diving. The focus of the camera shifts from the diving springboard to the water beneath as the action is executed. Moreover, the springboard might move as the person jumps into the water, and there may be other humans moving around the swimming pool. In the absence of any external localization ground-truth, the neural network needs to innately disentangle the human actor performing the action of diving from the rest of the scene.  In general, a video scene consists of four types of regions - dynamic salient regions, static salient regions, dynamic non-salient regions and static non-salient regions. The first two are very important for the underlying task of action recognition. A robust feature disentanglement technique should highlight pixels corresponding to static salient regions and dynamic salient regions and suppress the non-salient patches. 

The frequency domain of a signal is a powerful representation that can quantify the temporal change of spatial pixels, and hence can be used for predicting static and dynamic regions of the scene. To identify dynamic pixels, we can formulate the ``dynamic'' mask $M_{dynamic}$~\cite{kothandaraman2022fourier} as the dot product of the temporal FFT $F_{T}$ of the feature maps and the corresponding frequencies $f$, over the frequency spectrum $k=1....T$. For features $x$, mathematically,
\begin{equation}
     M_{FAR}(x) = \underbrace{\| {F_{T}}(x) \|_{2}^{2} \times \| f \|_{2}^{2}}_{M_{dynamic}} \odot x,
     \label{eq:far_fo}
\end{equation}
The formulation represented by Equation \ref{eq:far_fo} amplifies high amplitudes at high frequencies, and suppresses low amplitudes at low frequencies. The final dot product with the feature representation ensures that only salient moving regions are amplified, and moving background regions are suppressed. Since this equation only models dynamic regions of the scene, it causes inconsistencies in static parts of the scene corresponding to salient regions; it may also incorrectly disentangle dynamic backgrounds and noise. To alleviate this issue, we propose modeling the salient static parts of the video through a static mask $M_{static}$. 

High amplitudes at low frequencies correspond to static regions of the scene. Given a feature representation $x$, we mathematically model this relation as follows:
\begin{equation}
     M_{static} = \| {F_{T}}(x) \|_{2}^{2} \odot (1/(1+\| f \|_{2}^{2})),
     \label{eq:static_mask}
\end{equation}
The inverse shifted quadratic formulation of the frequency component ensures that the computation of the mask is stable. A dot product of $M_{static}$ and network features amplifies static, salient regions and heavily suppresses dynamic non-salient regions. $M_{static}$, in conjunction with the dynamic mask $M_{dynamic}$, forms a static-dynamic frequency mask prior $M$. Thus, the net mask encapsulating static, and dynamic parts of the scene is $M = M_{static} + M_{dynamic}$. $M$ guides the network to \textit{learn} object disentangled low-level feature representations, as described in the following section.

\subsection{Differentiable Disentanglement} 

Our next goal is to use the differentiable static-dynamic frequency masks to help the neural network learn disentangled feature representations. We guide the learning of disentangled feature representations at the penultimate layer of the network using the static-dynamic mask prior $M$ from mid-level features. Since mid-level features provide a fine balance between encoding video features and maintaining the space-time resolution of the video, we use mid-level features $x_{mid}$ to derive the disentanglement mask $x_{mid,M}$, computed by applying $M$ on $x_{mid}$. We constrain the features at the penultimate layer, $x_{p}$ to resemble disentangled feature representations using an identity loss function as follows:
\begin{equation}
     L_{mask} = \lambda_{mask} \cdot MSE \{ x_{p}, M(x_{mid}) \odot x_{p}\},
     \label{eq:disentangle_loss}
\end{equation}
where MSE is mean squared error. In the above equation, the dot product of mask $M(x_{mid})$ ($M$ applied on $x_{mid}$) and $x_{p}$ transforms $x_{p}$ to the corresponding disentangled feature representation. Further, the application of the identity loss enables the network to intrinsically learn high-quality disentangled feature representations at the penultimate layer. During backpropogation, gradients flow from $x_{p}$ as well as $x_{mid}$ (through $M$) ensuring that neural network layers between the mid-level and penultimate layer learn disentangled features. The hyperparameter $\lambda_{mask}$, typically set to $0.1$ in our benchmarks, scales the identity loss function. We use a combination of the classical multi-class cross entropy loss function \cite{carreira2017quo} applied on the final softmax layer of the neural network, along with $L_{mask}$, to train the action recognition neural network.

\subsection{Frame Sampling}

Uniform sampling divides the video into uniform segments, equal to the number of frames $N$ to be sampled, and selects the $k_{th}$  from each segment where $k$ is an integer lesser than or equal to the number of frames in each segment. Such a sampling strategy is commonly adopted during the training of action recognition architectures on trimmed videos and works well due to its ability to span the duration of the video. However, samples selected from each video segment using uniform sampling are not always the most salient frames in the corresponding segment. We present a method to choose the best frame within uniform video segments. Our solution can be applied at the evaluation phase and does not require the neural network to be retrained.

A good frame selection strategy should consider frames prior to the video segment under consideration as well as frames succeeding the video segment. Frames that provide maximal information w.r.t. prior as well as future frames can be regarded as salient frames. For the task of action recognition, movement or difference of feature representations between frames provides this information. Corresponding to segment $i$ ($i=1...T$), where $T$ is the number of frames, for frame $j$'s ($j=1...s$), where $s$ is the number of frames within each segment (or total number of frames in the video divided by sampling rate), we denote the feature representations $x_{i,j}$. Since movement is best encapsulated by the salient regions, which are useful for predictions, we compute the saliency mask representations for features $x_{i,j}$ as $M \odot x_{i,j}$. We denote uniform frames for segment $i$ as $x_{U-i}$, and the corresponding saliency mask representations as $M \odot x_{U-i}$. Weighted sum of mean squared difference between $M \odot x_{i,j}$ and prior and future frames gives the cost w.r.t. prior $C_{p}$ and future frames $C_{f}$. The weights are designed to be inversely proportional to the distance between the video segment under consideration for frame sampling and the video segment in the equation for computation for prior and future costs. Mathematically, 
\begin{equation}
     C_{p} = \sum_{k=1}^{i-1}\frac{T-i+k}{T} \times (M \odot x_{i,j} - M \odot x_{U-k})^{2} 
     \label{eq:sampling_prior}
\end{equation}
\begin{equation}
     C_{f} = \sum_{k=i+1}^{T}\frac{T+i-k}{T} \times(M \odot x_{i,j} - M \odot x_{U-k})^{2}. 
     \label{eq:sampling_future}
\end{equation}

The net cost is a weighted sum of $C_{p}$ and $C_{f}$ where the weights, $w_{p}$ and $w_{f}$, are proportional to the costs and time between the video segment under consideration for frame sampling, the beginning of the video, and the end of the video. Mathematically, $w_{p} = \{C_{p}/(C_{p} + C_{f}\} \times i$ and $w_{f} = \{C_{p}/(C_{p} + C_{f}\} \times (T-i)$. 

\paragraph{Ensembling the predictions.} We use the set of uniform frames $F_{U}$ to sample the set of salient frames $F_{S}$. We use the trained action recognition neural network to compute the predictions for $F_{U}$ as well as $F_{S}$. These predictions are summed to obtain the final prediction, which is used to determine the predicted action class. 
\section{Experiments}

\begin{table}
\centering 
\resizebox{0.95\columnwidth}{!}{
\begin{tabular}{ccccc}
\toprule
Model & Data Aug. & Input & Frames & Top-1 (\%)  \\
\midrule 
\multicolumn{5}{c}{I3D Experiments} \\
\midrule 
I3D & \xmark & ($540,960$) & $8$ & $23.86$\\
\rowcolor{RowColorCode}
Ours (I3D Backbone) & \xmark & ($540,960$) & $8$ & $32.94$ \\
\midrule 
\multicolumn{5}{c}{X3D Experiments} \\
\midrule 
X3D & \xmark & ($540,960$) & $16$ & $35.71$\\
\rowcolor{RowColorCode}
Ours (X3D backbone) & \xmark & ($540,960$) & $16$ & $40.81$ \\
\midrule 
\multicolumn{5}{c}{X3D + Data Augmentation Experiments} \\
\midrule 
X3D & \cmark & ($540,540$) & $8$ & $36.6$\\
\rowcolor{RowColorCode}
X3D + FD & \cmark & ($540,560$) & $8$ & $41.86$\\
\bottomrule
\end{tabular}
}
\caption{We evaluate our method on UAV Human. Across two 3D CNN backbones (I3D, X3D) and datasets, we demonstrate relative improvements of $14.28\% - 38.05\%$ in top-1 accuracies over the corresponding baselines.}
\label{tab:uavhuman_all}
\end{table}

\subsection{Network and Training Details}
\paragraph{Backbone network architecture:} We benchmark our models using two state-of-the-art 3D CNN based action recognition architectures :- I3D (CVPR 2017) and X3D (CVPR 2020). We use level $2$ features (mid-level) of the underlying 3D ResNet-101 to derive object disentanglement masks. This mask is used to guide the learning (using Equation \ref{eq:disentangle_loss}) of disentangled feature representations obtained after level $3$. The cost functions for frequency-based frame sampling are computed using features from level $1$. 

\paragraph{Training details:} All our models were trained using NVIDIA GeForce 1080 Ti GPUs and NVIDIA RTX A5000 GPUs. Initial learning rates were set at $0.01$ or $0.1$. We use the Stochastic Gradient Descent (SGD) for optimization, with $0.0005$ weight decay, and 0.9 momentum. We use cosine/poly annealying for learning rate decay and multi-class cross entropy loss to constrain the final softmax predictions of the neural network. We report top-1 accuracies for all our models. 

\paragraph{Datasets:} We benchmark our models on two UAV datasets:- UAV Human~\cite{li2021uav} and NEC Drone~\cite{choi2020unsupervised}. UAV Human is an outdoor activities dataset, containing low-resolution videos taken under adverse lighting and weather conditions. It contains $155$ actions, many of which are similar and hard to distinguish. Moreover, the videos contain dynamic backgrounds, camera motions and noise. NEC Drone is an indoor dataset with single-human and two-person actions. It contains $16$ actions performed in an unconstrained manner. 

\paragraph{Evaluation:} We evaluate all action recognition models using the top-1 accuracy score, which is the ratio of the number of correctly classified samples to the total number of samples in the evaluation set. We also report top-5 accuracy which is the ratio of samples for which the top-5 classes (in terms of probability of predictions) contains the ground-truth class. We report all improvements in terms of relative performance. The relative improvement of method 1 over method 2, where the absolute accuracies of method 1 and method 2 are $x_{1}\%$ and $x_{2}\%$ respectively, is $(x_{1} - x_{2})/ x_{2} \times 100$

\subsection{Results: UAV Human}

\begin{table}
\centering 
\resizebox{0.7\columnwidth}{!}{
\begin{tabular}{cc}
\toprule
Experiment & Top-1 (\%)  \\
\midrule 
$M_{dynamic}$ & $28.1$ \\ 
$M_{dynamic}$ + $L_{mask}$ & $29.68$ \\
$M_{static}$ + $M_{dynamic}$ + $L_{mask}$ & $31.95$ \\
\bottomrule
\end{tabular}
}
\caption{We demonstrate the effectiveness of each component of FD with ablation experiments on UAV Human. All our experiments use RIUS for training as well as testing, and I3D backbone, and a spatial and temporal resolution of $540 \times 960$ and $8$ frames respectively.}
\label{tab:fd_ablations}
\end{table}
\begin{table}
\centering 
\resizebox{0.99\columnwidth}{!}{
\begin{tabular}{cccccccc}
\toprule
Model & Backbone & Data Aug. & Input & Frames & Params (M) & FLOPs (G) & Top-1 (\%) \\
\midrule 
\multicolumn{7}{c}{UAV Human} \\
\midrule 
I3D (CVPR 2017) \cite{carreira2017quo}& - & \xmark & $(540,960)$ & $8$ & $28.0$ & $108$ & $21.06$\\
MVIT (ICCV 2021) \cite{fan2021multiscale} & - & \cmark & $(224,224)$ & $16$ & $36.6$ & $70.8$ & $24.3$ \\
X3D-M (CVPR 2020) \cite{feichtenhofer2020x3d} & - & \cmark & $(224,224)$ & $8$ & $3.8$ & $6.2$ & $30.6$ \\
X3D-M (CVPR 2020) \cite{feichtenhofer2020x3d}& - & \cmark & $(540,540)$ & $8$ & $3.8$ & $36.03$ & $36.6$ \\
X3D-M (CVPR 2020) \cite{feichtenhofer2020x3d}& - & \xmark & $(540,960)$ & $16$ & $3.8$ & $36.03$ & $x$ \\
FNet (arXiv 2021) \cite{fnet} & I3D & \xmark & $(540,960)$ & $8$ & $28.0$ & $108$ & $24.39$ \\
FAR (ECCV 2022) \cite{kothandaraman2022fourier} & I3D & \xmark & $(540,960)$ & $8$ & $28.0$ & $108$ & $29.21$\\
FAR (ECCV 2022) \cite{kothandaraman2022fourier} & X3D-M & \cmark & $(540,540)$ & $8$ & $3.8$ & $36.6$ & $38.6$ \\
\rowcolor{RowColorCode}
Ours & X3D-M & \cmark & $(540,540)$ & $8$ & $3.8$ & $36.6$ & $41.86$ \\
\rowcolor{RowColorCode}
Ours & X3D-M & \xmark & $(540,960)$ & $16$ & $3.8$ & $130.13$ & $40.81$ \\
\bottomrule
\end{tabular}
}
\caption{We compare our method against prior methods on UAV Human. We demonstrate state-of-the-art performance by atleast $5.72\%$. Our method outperforms transformer architectures while using far fewer computations (upto $61.8\times$) and network parameters (upto $31.94\times$). Our method imposes minimal memory overhead over the backbone neural network.}
\label{tab:uavhuman_sota}
\end{table}

\paragraph{Our method can be embedded within any state-of-the-art 3D CNN to improve the performance.} We demonstrate the effectiveness of our method over various backbone architectures in Table \ref{tab:uavhuman_all}. We report relative improvements of $38.05\%$ and $14.28\%$ using the I3D and X3D backbones respectively. The large improvements in top-1 accuracy are indicative of the ability of FD to inherently localize the human actor from the background and identify actions from obscured top-down and oblique views in the presence of dynamic UAV cameras and noise.

All models use uniform sampling (randomly initialized) \cite{kothandaraman2022fourier} for training. In the first two sets of experiments, we do not perform any data augmentation. The differentiable static-dynamic frequency mask and identity loss are used for training and the ensemble frequency-based frame sampling strategy is used at evaluation for our method. In the third set of experiments, the X3D model is trained with spatial data augmentations, consistent with the publicly available official code for X3D~\cite{feichtenhofer2020x3d} in Meta AI's SlowFast GitHub repository. While we use the differentiable static-dynamic frequency mask and identity loss for training, the frame sampling and evaluation scheme are consistent with the SlowFast repository codebase i.e. our ensemble frame selection strategy is not used for testing.

\paragraph{Ablation experiments on the differentiable static-dynamic frequency mask and identity loss.} We present ablation experiments on UAV Human in Table \ref{tab:fd_ablations}. We use the I3D backbone at spatial and temporal resolutions of $540\times 960$ and $8$ frames respectively. All experiments use uniform sampling (randomly initialized) for training as well as testing. 
\begin{itemize}
    \item In the first experiment, we apply only the dynamic mask $M_{dynamic}$ (Equation 1), as a transformation over mid-level features, as in FAR. The accuracy is 28.1\%.
    \cite{kothandaraman2022fourier}.
    \item In the second experiment, we quantitatively demonstrate the benefits of differentiability. We use $M_{dynamic}$ to train the network to learn disentangled features in a differentiable manner using $L_{mask}$ (Equation 3). The absolute improvement over the first experiment is $1.58\%$.
    \item In the third experiment, we show the effects of the static-dynamic mask. We use $M_{static}$ (Equation 2) along with the setup in the second experiment. The absolute improvement over the second experiment is $2.27\%$.
\end{itemize}

\begin{table}
\centering 
\resizebox{0.9\columnwidth}{!}{
\begin{tabular}{ccccc}
\toprule
Model & Input & Frames & Top-1 (\%) & Top-5 (\%)  \\
\midrule 
\multicolumn{5}{c}{I3D Experiments} \\
\midrule 
w/o sampling & ($540,960$) & $8$ & $31.75$ & $52.92$\\
with sampling & ($540,960$) & $8$ & $32.94$ & $54.4$\\
\midrule 
\multicolumn{5}{c}{X3D Experiments} \\
\midrule 
w/o sampling & ($540,960$) & $16$ & $40.12$ & $59.08$\\
with sampling & ($540,960$) & $16$ & $40.81$ & $59.94$\\
\bottomrule
\end{tabular}
}
\caption{We present ablation experiments on the frame sampling method. None of our models use data augmentation.}
\label{tab:uavhuman_sampling}
\end{table}

\paragraph{Effectiveness of the ensemble frame sampling method.} We present experiments on the effectiveness of the frame sampling method in Table \ref{tab:uavhuman_sampling}. On I3D, the frame sampling method improves top-1 and top-5 accuracies by (relative) $3.74\%$ and $2.79\%$ respectively. On X3D, the frame sampling method improves top-1 and top-5 accuracies by (relative) $1.71\%$ and $1.45\%$ respectively. 

\paragraph{State-of-the-art comparisons.} We present comparisons against the state-of-the-art in Table \ref{tab:uavhuman_sota}. We compare against 3D CNNs (I3D \cite{carreira2017quo}, X3D \cite{feichtenhofer2020x3d}), transformer models (MVIT \cite{fan2021multiscale}) and frequency-based methods (FNet \cite{fnet}, FAR \cite{kothandaraman2022fourier}). Our method has benefits in terms of top-1 accuracy as well as computation (FLOPs, number of parameters). It has minimal memory overheads over the corresponding backbone neural network. We demonstrate (relative) improvements of atleast {\bf $5.72\%$} over the state-of-the-art method.  

\paragraph{Ablation experiments on the frame sampling method.} All experiments are conducted on the X3D backbone, without data augmentation and a spatial-temporal resolution of $16 \times 540 \times 960$. The purpose of the disentanglement mask in frame sampling is to ensure samples are chosen based on the relevance of only salient features. When applied on a model trained for learning disentangled features, a good way to verify if the network has actually learnt disentangled features is to remove the mask in the calculation of cost functions for frame sampling. This results in an accuracy of $40.9\%$, proving that our differentiable method is indeed successful. Setting $w_{p} = w_{f} = 1$ results in an accuracy of $40.70\%$.  

\begin{table}
\centering 
\resizebox{0.99\columnwidth}{!}{
\begin{tabular}{ccccccc}
\toprule
Model & Backbone & Input & Frames & Params (M) & FLOPs (G) & Top-1 (\%) \\
\midrule 
X3D-M (CVPR 2020) \cite{feichtenhofer2020x3d}& - & $(960,540)$ & $8$ & $3.8$ & $64.05$ & $66.15$ \\
FAR (ECCV 2022) \cite{kothandaraman2022fourier} & X3D-M & $(960,540)$ & $8$ & $3.8$ & $65.08$ & $71.46$ \\
\rowcolor{RowColorCode}
Ours & X3D-M & $(960,540)$ & $8$ & $3.8$ & $65.08$ & $80.75$ \\
\bottomrule
\end{tabular}
}
\caption{We present experiments on NEC Drone. We demonstrate improvements of $13\% - 22\%$ over prior work.}
\label{tab:nec_drone}
\end{table}

\subsection{Results: NEC Drone}

We present results on NEC-Drone in Table \ref{tab:nec_drone}. We use uniform sampling~\cite{kothandaraman2022fourier} to train all models. In the experiment corresponding to our method, we initialize the neural network with weights corresponding UAV Human. Following prior work~\cite{kothandaraman2022fourier}, while testing, we select the most frequent class prediction among the networks' result using all sets of Randomly Initialized Uniform Sampling frames.

We obtain a top-1 accuracy of $80.75\%$. We compare against 3D CNNs (X3D~\cite{feichtenhofer2020x3d}) and frequency-based methods (current state-of-the-art for aerial video recognition - FAR~\cite{kothandaraman2022fourier}) and show relative improvements of {\bf 13.00\% - 22.07\%}. 
\section{Conclusion, LImitations and Future Work}

In this paper, we presented a differentiable disentanglement method that simultaneously exploits the potential of learning-based neural networks and traditional signal processing. The method includes a static-dynamic frequency mask and identity loss for learning the disentanglement, along with a frame selection method at the evaluation stage. Our method has a few limitations. First, it is completely supervised. This means it requires labeled data and annotation is a laborious process. Unsupervised, self-supervised and transfer learning methods can alleviate this issue. Second, while our method is more effective than transformers in terms of accuracy, speed and memory, it is still expensive due to 3D convolutions and requires multiple GPUs and iterations to train. Future work on methods that can be run on mobile GPUs will be of practical importance. In our experiments and datasets, we assumed that a single human-agent is performing the action. Our method can be incorporated within any 3D neural network to improve performance and it might be interesting to explore the usage of our method within multi-agent neural networks.

\noindent \textbf{Acknowledgements:} This research has been supported by ARO Grants W911NF2110026 and Army Cooperative Agreement W911NF2120076




\bibliographystyle{IEEEtran}
\bibliography{references}




\end{document}